\documentclass[conference]{IEEEtran}
\IEEEoverridecommandlockouts

\usepackage{amsmath,amssymb}
\usepackage{graphicx}
\usepackage{cite}
\usepackage{booktabs}
\usepackage{algorithm}
\usepackage{algpseudocode}

\title{Towards Noise-Resilient Quantum Multi-Armed and Stochastic Linear Bandits}

\author{
\IEEEauthorblockN{Zhuoyue~Chen}
\IEEEauthorblockA{School of
Electronics and Communication Engineering\\
Sun Yat-sen
University\\
Shenzhen, China\\
chenzhy225@mail2.sysu.edu.cn}
\and
\IEEEauthorblockN{Kechao~Cai}
\IEEEauthorblockA{School of
Electronics and Communication Engineering\\
Sun Yat-sen
University\\
Shenzhen, China\\
caikch3@mail.sysu.edu.cn}
}

\begin{document}

\maketitle

\begin{abstract}
Quantum multi-armed bandits (MAB) and stochastic linear bandits (SLB) have recently attracted significant attention, as their quantum counterparts can achieve quadratic speedups over classical MAB and SLB. However, most existing quantum MAB algorithms assume ideal quantum Monte Carlo (QMC) procedures on noise-free circuits, overlooking the impact of noise in current noisy intermediate-scale quantum (NISQ) devices. In this paper, we study a noise-robust QMC algorithm that improves estimation accuracy when querying quantum reward oracles. Building on this estimator, we propose noise-robust QMAB and QSLB algorithms that enhance performance in noisy environments while preserving the advantage over classical methods. Experiments show that our noise-robust approach improves QMAB estimation accuracy and reduces regret under several quantum noise models.
\end{abstract}


\section{Introduction}
Quantum computing has demonstrated significant potential to accelerate computational tasks that are classically intractable, particularly in sampling and high-dimensional estimation problems. Among these, quantum Monte Carlo (QMC)~\cite{montanaro2015quantum} methods provide a quadratic speedup over classical Monte Carlo by reducing the sampling complexity from $O(1/\epsilon^2)$ to $O(1/\epsilon)$, where $\epsilon$ denotes the target estimation error.
Such improvements are especially appealing in learning and optimization scenarios where repeated expectation estimation is required.

Online decision-making problems, notably multi-armed bandits (MAB)~\cite{auer2002finite} and stochastic linear bandits (SLB)~\cite{li2010contextual}, constitute a fundamental class of sequential learning tasks with wide applications in recommendation systems~\cite{zhou2015survey}, wireless communications~\cite{bande2019multi}, online advertising~\cite{graepel2010web}, and adaptive control~\cite{borkar2008stochastic}. In these problems, efficient estimation of reward distributions directly determines the regret performance and long-term utility. Consequently, accelerating reward estimation can substantially improve learning efficiency and decision quality. Quantum computing can accelerate sampling and reward estimation, which motivates the development of quantum multi-armed bandit (QMAB) algorithms that leverage QMC-based sampling advantages.

Despite these theoretical benefits, practical quantum devices currently operate in the noisy intermediate-scale quantum (NISQ) regime~\cite{preskill2018quantum}, where decoherence, gate errors, and measurement noise significantly degrade the sampling accuracy of QMC estimators. In noisy environments, QMC estimators may suffer from variance amplification and bias accumulation, potentially eliminating their theoretical speedup~\cite{suzuki2020amplitude,aaronson2020quantum}. Existing quantum bandit algorithms typically assume ideal, noise-free QMC and overlook the impact of circuit-level noise~\cite{casale2020quantum,wan2023quantum}; consequently, their performance can degrade on real devices, reducing their advantage over classical baselines. As a result, the practical realization of regret-optimal quantum bandits under realistic hardware conditions remains largely unexplored.

While prior works have investigated quantum speedups for Monte Carlo estimation and have proposed quantum algorithms for MAB and SLB, robustness under NISQ noise has received limited attention. Existing approaches~\cite{wan2023quantum,casale2020quantum} typically focus on idealized noise-free settings or treat error mitigation techniques independently from online learning dynamics. In practical settings, noise can make QMC estimates inaccurate and, in turn, degrade the performance of quantum bandit algorithms. Consequently, a unified framework that integrates noise-resilient QMC estimation with quantum bandit algorithms remains lacking. 

In this paper, we propose a noise-resilient quantum Monte Carlo framework tailored for online learning. We first develop a noise-robust QMC algorithm based on Bayesian estimation, BQMC, which can improve estimation accuracy in noisy settings. Building upon BQMC, we develop noise-resilient quantum multi-armed and stochastic linear bandit algorithms, including noise-resilient quantum UCB (NR-QUCB) and noise-resilient quantum LinUCB (NR-QLinUCB). The proposed framework preserves logarithmic regret behavior while maintaining robustness under realistic noise assumptions. Extensive simulations under serveral noise models validate the effectiveness of the proposed approach.

The main contributions of this paper are summarized as follows:
\begin{itemize}
    \item We propose a noise-robust QMC algorithm based on Bayesian estimation, BQMC, which improves estimation performance in noisy environments.
    \item We integrate the BQMC algorithm with quantum bandit methods and develop noise-resilient quantum UCB (NR-QUCB) and noise-resilient quantum LinUCB (NR-QLinUCB), improving regret performance under realistic noise.
    \item We conduct extensive experiments to demonstrate that our quantum bandit algorithms achieve improved regret performance across multiple noise models.
\end{itemize}

\section{Background}
\subsection{Quantum Bandits}
A quantum system is described by a state vector in a complex Hilbert space $\mathcal{H}=\mathbb{C}^{2^n}$. An $n$-qubit pure state can be written as
\begin{equation}
    \lvert \psi \rangle = \sum_{x \in \{0,1\}^{n}} \alpha_{x} \lvert x \rangle,
\end{equation}
where $\sum_{x} \lvert \alpha_{x} \rvert^{2} = 1$, and measurement returns outcome $x$ with probability $\lvert \alpha_{x} \rvert^{2}$.
Quantum algorithms exploit superposition and interference to manipulate probability amplitudes coherently, enabling potential computational speedups over classical randomized procedures.

In the stochastic multi-armed bandit problem, we are given a finite action set $\mathcal{X} = \{1,\ldots,K\}$, where each arm $x \in \mathcal{X}$ is associated with an unknown reward distribution $\nu_x$, and pulling arm $x$ produces a random reward $r_x \in \{0,1\}$ sampled from $\nu_x$ with mean $a_x = \mathbb{E}[r_x]$. The goal is to identify the optimal arm $x^{*} = \arg\max_{x \in \mathcal{X}} a_x$ while minimizing cumulative regret. In classical algorithms such as upper confidence bound (UCB), the empirical mean is estimated via repeated sampling, and achieving precision $\epsilon$ requires $O(1/\epsilon^{2})$ samples per arm.
In the quantum setting, we adopt a black-box model where each action $x$ is associated with a quantum reward oracle $O_x$, and the objective is to estimate the mean reward $a_x = \mathbb{E}[r_x]$, where $r_x$ denotes the reward associated with arm $x$, while identifying the optimal arm $x^{*} = \arg\max_{x \in \mathcal{X}} a_x$. Thus, quantum multi-armed bandit consists of a family of unitary oracles $\{O_x\}_{x \in \mathcal{X}}$, and the learning task reduces to efficiently estimating the expectation encoded in each oracle while minimizing the total number of oracle queries.
The total number of oracle queries is defined as
\begin{equation}
    Q = \sum_{x \in \mathcal{X}} Q_x,
\end{equation}
where $Q_x$ denotes the number of queries to oracle $O_x$. This formulation mirrors the classical setting, where each arm corresponds to a distinct stochastic process, but now each arm is a quantum subroutine.

\subsection{Quantum Reward Oracles}
A central component of quantum bandit is the quantum reward oracle, which encodes reward information into a unitary transformation.

In the quantum bandit, each arm $x \in \mathcal{X}$ is modeled as an independent quantum reward oracle. Specifically, we assume access to a family of unitary operators $\{O_x\}_{x \in \mathcal{X}}$, where oracle $O_x$ encodes the mean reward $a_x \in [0,1]$ of arm $x$ into a quantum amplitude. A canonical realization of such an oracle acts on an ancilla qubit as
\begin{equation}
    O_x \lvert 0 \rangle = \sqrt{1-a_x}\,\lvert 0 \rangle + \sqrt{a_x}\,\lvert 1 \rangle,
\end{equation}
so that measuring the output qubit in the computational basis yields outcome $1$ with probability $a_x$.
More generally, the oracle may act on a larger Hilbert space, preparing a joint state
\begin{equation}
    O_x \lvert 0 \rangle \lvert 0 \rangle = \sum_{y} \sqrt{\nu_x(y)}\,\lvert y \rangle \lvert r(x,y) \rangle,
\end{equation}
where $\nu_x$ denotes the reward distribution of arm $x$ and $r(x,y) \in \{0,1\}$ is the reward function. 

Under this oracle model, estimating the mean reward of arm $x$ reduces to estimating the amplitude of state $\lvert 1 \rangle$ in the output of $O_x$. This task can be accomplished using quantum amplitude estimation (QAE), which achieves precision $\epsilon$ using $O(1/\epsilon)$ oracle queries, compared with the classical $O(1/\epsilon^2)$ sampling complexity. The total complexity of a quantum bandit algorithm is therefore measured by the total number of oracle calls across all arms, and the quadratic improvement in precision directly translates into potential reductions in exploration cost. However, standard quantum amplitude estimation assumes coherent and noise-free oracle access, an assumption that is generally violated in NISQ devices. Consequently, investigating how noise affects the oracle amplitudes and designing robust estimation procedures becomes essential for practical quantum bandit algorithms.

\subsection{Quantum Monte Carlo}
The estimation of mean rewards in bandit algorithms can be accelerated via quantum Monte Carlo techniques~\cite{montanaro2015quantum}. In the oracle model described above, each arm $x$ prepares a quantum state of the form $\lvert \psi_x \rangle = \sqrt{1-a_x}\lvert 0 \rangle + \sqrt{a_x}\lvert 1 \rangle$. Estimating $a_x$ therefore reduces to estimating the amplitude of the marked state $\lvert 1 \rangle$. Quantum amplitude estimation (QAE) provides a systematic procedure to approximate this amplitude with precision $\epsilon$ using $O(1/\epsilon)$ oracle queries. More generally, if the reward-generating process is represented by a quantum subroutine whose output has bounded variance $\sigma^2$, quantum mean estimation algorithms can approximate the expectation with precision $\epsilon$ using $O(\sigma/\epsilon)$ oracle invocations. This improvement originates from amplitude amplification, where probability amplitudes grow linearly with the number of iterations, yielding a quadratic enhancement in precision scaling. Nevertheless, practical implementations on NISQ hardware introduce noise into oracle realizations, which perturbs amplitudes and may bias the estimation process. Consequently, developing noise-resilient quantum Monte Carlo estimators is essential for maintaining quantum acceleration under realistic conditions.

\section{System Model}
\subsection{Noisy Quantum Bandit Model}
We consider a stochastic quantum multi-armed bandit with action set $\mathcal{X} = \{1,\ldots,K\}$.
Each arm $x \in \mathcal{X}$ is associated with an unknown mean reward $a_x \in [0,1]$, and the objective is to identify and exploit the optimal arm $ x^{*} = \arg\max_{x \in \mathcal{X}} a_x $.
In the ideal quantum setting, each arm is modeled as a quantum reward oracle $O_x$ that prepares a state
\begin{equation}
    \lvert \psi_x \rangle = \sqrt{1-a_x}\lvert 0 \rangle + \sqrt{a_x}\lvert 1 \rangle .
\end{equation}
Estimating the mean reward $a_x$ reduces to estimating the amplitude of the marked state $\lvert 1 \rangle$.

On realistic NISQ devices, oracle implementations are imperfect. Instead of directly accessing the ideal unitary $O_x$, we assume that the implemented oracle is a noisy quantum operation described by
\begin{equation}
    \tilde{O}_x = \mathcal{E}_x \circ \mathcal{U}_x,
\end{equation}
where $\mathcal{U}_x(\rho) = O_x \rho O_x^{\dagger}$ denotes the unitary quantum channel corresponding to the ideal oracle, $\mathcal{E}_x$ is a completely positive trace-preserving (CPTP) noise channel modeling hardware imperfections, and the symbol ``$\circ$'' denotes composition of quantum channels, i.e., $(\mathcal{E}_x \circ \mathcal{U}_x)(\rho) = \mathcal{E}_x(\mathcal{U}_x(\rho))$.
Thus, the implemented oracle first applies the ideal unitary transformation and subsequently undergoes a noise process characterized by $\mathcal{E}_x$.
Let $\rho_0 = \lvert 0 \rangle \langle 0 \rvert$ denote the initial state. The noisy output state is
\begin{equation}
    \tilde{\rho}_x = \tilde{O}_x(\rho_0) = \mathcal{E}_x(O_x \rho_0 O_x^{\dagger}).
\end{equation}
We define the noisy reward as $\tilde{a}_x = \mathrm{Tr}(\lvert 1 \rangle \langle 1 \rvert \tilde{\rho}_x)$.
In general, $\tilde{a}_x \neq a_x$ due to the action of the noise channel.

To facilitate theoretical analysis, we adopt an effective statistical abstraction of the noisy reward probability. We model $\tilde{a}_x = a_x + b_x + \xi_x$,
where $b_x$ denotes a deterministic bias induced by systematic or coherent noise, and $\xi_x$ denotes a stochastic fluctuation capturing random noise effects.
We assume $\mathbb{E}[\xi_x] = 0, \mathrm{Var}(\xi_x) \le \sigma_x^2$,
and $\lvert b_x \rvert \le \delta$,
for some known bounds $\sigma_x$ and $\delta$.
This decomposition separates systematic distortion from random uncertainty and provides a unified framework for analyzing noise-robust estimation procedures. Under this model, the learner interacts with noisy oracle outputs governed by $\tilde{a}_x$, while regret is evaluated with respect to the true means $a_x$.

\subsection{Performance Metrics}
\textbf{Cumulative Regret.} Let $x_t$ denote the arm selected at round $t$. The cumulative regret after $T$ rounds is defined as
\begin{equation}
    R_T = \sum_{t=1}^{T} \bigl(a_{x^{*}} - a_{x_t}\bigr).
\end{equation}
Regret is measured with respect to the true mean rewards $a_x$. Noisy oracle perturbations may lead to suboptimal arm selection and thus increase regret. A noise-resilient algorithm aims to control the growth of $R_T$ despite biased and stochastic reward observations.


\section{Algorithm}
\subsection{Bayesian Quantum Monte Carlo}






Estimating the mean reward encoded in a quantum oracle can be formulated as an amplitude estimation problem. In this subsection, we derive a Bayesian Quantum Monte Carlo (BQMC) estimator that is robust under noisy oracles.

Consider an arm $x$ associated with a reward oracle $O_x$. Let the ideal state preparation unitary produce
\begin{equation}
    O_x \lvert 0 \rangle = \sin \theta_x \lvert \psi_1 \rangle + \cos \theta_x \lvert \psi_0 \rangle,
\end{equation}
where the probability of observing reward $1$ equals $a_x = \sin^2 \theta_x$.
Applying the Grover operator $Q$ repeatedly yields
\begin{equation}
    Q^{m} \lvert \psi \rangle = \sin((2m + 1)\theta_x)\lvert \psi_1 \rangle + \cos((2m + 1)\theta_x)\lvert \psi_0 \rangle.
\end{equation}
Hence, measuring the output qubit after depth $m$ gives
\begin{equation}
    p(\theta_x, m) = \sin^2((2m + 1)\theta_x).
\end{equation}
This nonlinear transformation amplifies sensitivity and underlies the quadratic speedup principle.

Under the noisy oracle model $\tilde{O}_x = \mathcal{E}_x \circ O_x$,
the success probability becomes distorted. We denote the noisy success probability as $\tilde{p}(\theta_x,m)$. A typical effective model under depolarizing noise is
\begin{equation}
    \tilde{p}(\theta_x,m) = (1-\eta)\sin^2((2m+1)\theta_x) + \frac{\eta}{2},
\end{equation}
where $\eta$ is the noise level. With $S$ shots and $h$ observed successes, the likelihood is binomial:
\begin{equation}
    \mathcal{L}(\theta_x) = \binom{S}{h}\,\tilde{p}(\theta_x,m)^{h}\bigl(1-\tilde{p}(\theta_x,m)\bigr)^{S-h}.
\end{equation}

Inspired by~\cite{ramoa2025bayesian}, instead of estimating $a_x$ directly, we perform Bayesian inference over $\theta_x$.
We assume $\theta_x \sim \mathcal{U}(0,\pi/2)$.
This corresponds to a non-informative prior over amplitudes, since $a_x = \sin^2 \theta_x$.
Let $\pi(\theta_x)$ denote the prior probability density function of $\theta_x$.
Given $h$, Bayes' rule gives
\begin{equation}
    \pi(\theta_x \mid h) \propto \pi(\theta_x)\tilde{p}(\theta_x,m)^{h}\bigl(1-\tilde{p}(\theta_x,m)\bigr)^{S-h}.
\end{equation}
Because conjugacy is lost due to the nonlinear sine transformation, we approximate the posterior using Sequential Monte Carlo (SMC)~\cite{smith2013sequential,del2006sequential,south2019sequential}.

We represent the posterior by particles: $\{\theta^{(i)}, w^{(i)}\}_{i=1}^{M}$, where $w^{(i)}$ denotes the normalized importance weight associated with particle $\theta^{(i)}$.
The posterior expectation of the reward amplitude is then
\begin{equation}
    \hat{a}_x = \mathbb{E}[a_x] = \sum_{i=1}^{M} w^{(i)} \sin^2 \theta^{(i)} .
\end{equation}
The posterior standard deviation is
\begin{equation}
    s_x = \sqrt{\sum_{i=1}^{M} w^{(i)} \bigl(\sin^2 \theta^{(i)} - \hat{a}_x \bigr)^2 } .
\end{equation}
This variance estimate will later serve as the exploration term in QUCB and QLinUCB.

Algorithm~\ref{alg:BQMC} presents the detailed procedure of Bayesian Quantum Monte Carlo (BQMC) for estimating the mean reward amplitude associated with a noisy quantum reward oracle. 
Given a noisy oracle $\tilde{\mathcal{O}}_x$, the algorithm performs $N$ Bayesian update rounds, where in each round the oracle is queried using $S$ shots at an adaptively selected amplification depth.
The algorithm takes as input the noisy oracle $\tilde{\mathcal{O}}_x$, the number of Bayesian rounds $N$, the number of shots per round $S$, the number of particles $M$, and the maximum allowable amplification depth $m_{\max}$. 
The output consists of the posterior mean estimate $\hat{a}_x$ and its posterior standard deviation $s_x$.

The posterior distribution of the unknown amplitude parameter $\theta_x$ is approximated by a set of weighted particles $\{\theta^{(i)}, w^{(i)}\}_{i=1}^{M}$. 
The particles are initialized by sampling $\theta^{(i)}$ uniformly over $[0,\pi/2]$ with equal weights (Line~\ref{alg:bqmc:init}). 
This corresponds to a non-informative prior over the amplitude parameter, where the reward mean satisfies $a^{(i)}=\sin^2\theta^{(i)}$.
In each Bayesian round $t$ (Line~\ref{alg:bqmc:for-start}), the algorithm first selects an amplification depth $m_\star$ that minimizes the expected posterior variance (Line~\ref{alg:bqmc:depth}). 
This adaptive depth selection balances information gain and noise amplification under the noisy oracle model.
The noisy oracle $\tilde{\mathcal{O}}_x$ is then executed for $S$ shots at depth $m_\star$ (Lines~\ref{alg:bqmc:oracle-start}–\ref{alg:bqmc:oracle-end}), producing $h$ successful outcomes. 
For each particle $\theta^{(i)}$, the likelihood $\tilde p(\theta^{(i)},m_\star)$ is computed (Line~\ref{alg:bqmc:likelihood}), and the importance weight $w^{(i)}$ is updated according to Bayes' rule (Line~\ref{alg:bqmc:update-weight}). 
The weights are subsequently normalized (Line~\ref{alg:bqmc:normalize}).
To prevent particle degeneracy, the effective sample size (ESS) is monitored. 
If the ESS falls below half of the particle number, resampling is performed (Line~\ref{alg:bqmc:resample}), which improves numerical stability and preserves posterior diversity.
After completing $N$ rounds, the posterior mean estimate $\hat{a}_x$ is computed as the weighted empirical expectation of $\sin^2\theta^{(i)}$ (Line~\ref{alg:bqmc:mean}), and the posterior standard deviation $s_x$ is obtained accordingly (Line~\ref{alg:bqmc:std}). 
The variance estimate $s_x$ will later serve as the exploration term in QUCB and QLinUCB.

\begin{algorithm}[t]
\caption{Bayesian Quantum Monte Carlo (BQMC)}
\label{alg:BQMC}
\begin{algorithmic}[1]
\Statex \textbf{Input:} Noisy oracle $\tilde{\mathcal{O}}_x$, rounds $N$, shots $S$, particles $M$, max depth $m_{\max}$
\Statex \textbf{Output:} Posterior mean estimate $\hat{a}_x$ and posterior standard deviation $s_x$

\State \label{alg:bqmc:init}
Sample $\theta^{(i)} \sim \mathcal{U}(0,\pi/2)$, set $w^{(i)} \leftarrow 1/M$

\For{$t=1$ to $N$}
    \label{alg:bqmc:for-start}
    \State $a^{(i)}\gets\sin^2\theta^{(i)}$
    \State \label{alg:bqmc:depth}
    $m_\star \leftarrow \arg\min_{0\le m\le m_{\max}}
    \mathbb{E}\!\left[\mathrm{Var}(a)\mid m\right]$
    
    \State \label{alg:bqmc:oracle-start}
    Prepare initial state $\lvert 0 \rangle$
    
    \State Apply noisy oracle $\tilde{\mathcal{O}}_x$ of depth $m_\star$ for $S$ shots
    \label{alg:bqmc:oracle-end}
    
    \State Measure output qubit and let $h$ be the number of outcomes $1$
    
    \For{$i=1$ to $M$}
        \State \label{alg:bqmc:likelihood}
        $p_i \leftarrow \tilde{p}(y=1\mid\theta^{(i)},m_\star)$
        
        \State \label{alg:bqmc:update-weight}
        $w^{(i)} \leftarrow w^{(i)} p_i^{\,h}(1-p_i)^{S-h}$
    \EndFor
    
    \State \label{alg:bqmc:normalize}
    Normalize $\{w^{(i)}\}$
    
    \State \label{alg:bqmc:resample}
    \textbf{if} $\mathrm{ESS}(\{w^{(i)}\}) < 0.5M$ \textbf{then resample}
    
\EndFor

\State \label{alg:bqmc:mean}
$\hat{a}_x \leftarrow \sum_{i=1}^{M} w^{(i)} \sin^2(\theta^{(i)})$

\State \label{alg:bqmc:std}
$s_x \leftarrow \sqrt{\sum_{i=1}^{M} w^{(i)}\big(\sin^2(\theta^{(i)})-\hat{a}_x\big)^2}$

\State \textbf{return} $(\hat{a}_x, s_x)$
\end{algorithmic}
\end{algorithm}

\subsection{Noise-Resilient Quantum UCB}

Existing Quantum UCB (QUCB) algorithms rely on quantum Monte Carlo to estimate the reward of each arm and construct upper confidence bounds accordingly. 
However, these algorithms are typically analyzed under idealized assumptions, where the reward oracle is noiseless and the amplitude estimation procedure is unbiased. 
In practical NISQ devices, oracle implementations are subject to stochastic and coherent noise, which distort the amplified amplitudes and introduce bias and additional variance into the reward estimates. 
As a result, the confidence bounds constructed from ideal QMC outputs may no longer be statistically valid, leading to unstable arm selection and degraded regret performance.

To address this limitation, we propose a Noise-Resilient Quantum UCB (NR-QUCB) algorithm that organically integrates the Bayesian Quantum Monte Carlo (BQMC) estimator with the UCB decision principle. 
Instead of relying on point estimates obtained from idealized QMC procedures, NR-QUCB employs the posterior mean and posterior variance returned by Algorithm~\ref{alg:BQMC} as noise-aware reward estimates and uncertainty measures. 
By embedding Bayesian inference in quantum reward estimation, NR-QUCB improves noise robustness while preserving UCB's optimism under uncertainty.

Algorithm~\ref{alg:NRQUCB} presents our \emph{Noise-Resilient Quantum UCB} (NR-QUCB) algorithm for noisy quantum multi-armed bandits. NR-QUCB follows the standard UCB principle by maintaining, for each arm $x \in \mathcal{X}$, (i) the current reward estimate $\hat{a}_x$, (ii) an uncertainty measure $s_x$, and (iii) a pull counter $n_x$. This mirrors the QUCB design~\cite{wan2023quantum}, where the learner keeps an estimate and a confidence radius for each arm and selects the arm maximizing the upper confidence bound.

Different from the stage-based QUCB construction, our NR-QUCB directly integrates the proposed Bayesian QMC as a noise-resilient mean estimator. Concretely, we first set a per-round failure probability $\delta_r = \delta/\max(1,T)$ and define a Gaussian-style exploration multiplier $z = \sqrt{2\log(T/\delta_r)}$ (Lines~\ref{alg:nrqucb:init}--\ref{alg:nrqucb:z}).
At each round $t$, NR-QUCB computes an index $I_x$ for every arm (Lines~\ref{alg:nrqucb:compute-index-start}--\ref{alg:nrqucb:compute-index-end}). For arms that have never been pulled ($n_x=0$), we set the upper confidence bound $I_x=+\infty$ to enforce at least one exploration (Lines~\ref{alg:nrqucb:ifzero}--\ref{alg:nrqucb:inf}). Otherwise, we use the upper confidence bound $I_x = \hat{a}_x + z\, s_x$,
where $\hat{a}_x$ and $s_x$ are returned by Algorithm~\ref{alg:BQMC} as the posterior mean and posterior standard deviation, respectively (Line~\ref{alg:nrqucb:index}). NR-QUCB then selects $x_t=\arg\max_x I_x$ (Line~\ref{alg:nrqucb:select}) and invokes Algorithm~\ref{alg:BQMC} on the corresponding noisy reward oracle $\tilde{\mathcal{O}}_{x_t}$ to refresh $(\hat{a}_{x_t}, s_{x_t})$ (Line~\ref{alg:nrqucb:bqmc}), followed by updating the counter $n_{x_t} \leftarrow n_{x_t}+1$ (Line~\ref{alg:nrqucb:update-n}). By coupling UCB optimism with noise-resilient Bayesian estimation, NR-QUCB yields a simple yet robust quantum bandit algorithm for NISQ settings while remaining within the UCB framework adopted by QUCB.

\begin{algorithm}[t]
\caption{Noise-Resilient Quantum UCB (NR-QUCB)}
\label{alg:NRQUCB}
\begin{algorithmic}[1]
\Statex \textbf{Input:} Arms $\mathcal{X}$, time horizon $T$, confidence level $\delta$
\Statex \textbf{Output:} Arm selection sequence $\{x_t\}_{t=1}^T$

\State Initialize $n_x \leftarrow 0$, $\hat{a}_x \leftarrow 0$, $s_x \leftarrow 0$, $I_x \leftarrow 0$ for all $x\in\mathcal{X}$
\label{alg:nrqucb:init}
\State $\delta_{\mathrm{r}} \leftarrow \delta/\max(1,T)$
\label{alg:nrqucb:deltar}
\State $z \leftarrow \sqrt{2 \log\!\left(\frac{T}{\delta_{\mathrm{r}}}\right)}$
\label{alg:nrqucb:z}

\For{$t = 1$ to $T$}
    \For{each arm $x \in \mathcal{X}$}
        \label{alg:nrqucb:compute-index-start}
        \If{$n_x = 0$}
            \label{alg:nrqucb:ifzero}
            \State $I_x \leftarrow +\infty$
            \label{alg:nrqucb:inf}
        \Else
            \State $I_x \leftarrow \hat{a}_x + z\, s_x$
            \label{alg:nrqucb:index}
        \EndIf
    \EndFor
    \label{alg:nrqucb:compute-index-end}

    \State $x_t \leftarrow \arg\max_{x\in\mathcal{X}} I_x$
    \label{alg:nrqucb:select}

    \State $(\hat{a}_{x_t}, s_{x_t}) \leftarrow \textbf{BQMC}(\tilde{\mathcal{O}}_{x_t})$
    \label{alg:nrqucb:bqmc}
    \State $n_{x_t} \leftarrow n_{x_t} + 1$
    \label{alg:nrqucb:update-n}
\EndFor

\State \textbf{return} $\{x_t\}_{t=1}^T$
\end{algorithmic}
\end{algorithm}

\subsection{Noise-Resilient Quantum LinUCB}

While NR-QUCB addresses stochastic quantum bandits with independent arms, many practical problems exhibit linear reward model. 
In quantum linear bandits, the expected reward of arm $x$ is assumed to satisfy
$
\mathbb{E}[r_x] = \phi_x^\top \theta^\star,
$
where $\phi_x \in \mathbb{R}^d$ is a known feature vector and $\theta^\star$ is an unknown parameter. 
Classical LinUCB maintains a ridge regression estimator of $\theta^\star$ and selects arms according to an optimism-based confidence ellipsoid.
Existing quantum LinUCB variants typically assume accurate quantum reward estimation. 
However, under noisy oracle implementations, the estimated rewards may exhibit bias and inflated variance, which propagate into the regression estimate and compromise the confidence bound. 
To mitigate this issue, we integrate the Bayesian Quantum Monte Carlo (BQMC) estimator into linear UCB, resulting in the proposed Noise-Resilient Quantum LinUCB (NR-QLinUCB).

Algorithm~\ref{alg:NRQLinUCB} outlines the detailed procedure of NR-QLinUCB. 
The algorithm takes as input the feature vectors $\{\phi_x\}_{x\in\mathcal{X}}$, the time horizon $T$, the regularization parameter $\lambda$, and the confidence level $\delta$. 
For each arm, it maintains a noise-aware reward estimate $\hat{a}_x$ and posterior uncertainty $s_x$, obtained via BQMC.
The algorithm initializes the ridge regression matrix $V=\lambda I$ and vector $b=0$ (Lines~\ref{alg:nrqlin:initV}). 
At each round $t$, the parameter estimate is computed as 
$
\hat{\theta}=V^{-1}b
$
(Line~\ref{alg:nrqlin:theta}).
For every arm $x$, an optimistic index is constructed (Lines~\ref{alg:nrqlin:index-start}--\ref{alg:nrqlin:index-end}) as
\[
I_x = \phi_x^\top \hat{\theta}
+ \sqrt{\beta_t\, \phi_x^\top V^{-1}\phi_x}
+ z\, s_x,
\]
where the second term corresponds to the classical LinUCB confidence ellipsoid, and the third term incorporates the posterior standard deviation returned by BQMC. 
This additional term explicitly accounts for noise-induced uncertainty in quantum reward estimation.
The arm $x_t=\arg\max_x I_x$ is selected (Line~\ref{alg:nrqlin:select}), and BQMC is invoked on the corresponding noisy oracle $\tilde{\mathcal O}_{x_t}$ to update $(\hat a_{x_t}, s_{x_t})$ (Line~\ref{alg:nrqlin:bqmc}). 
The regression statistics are then updated as
$
V \leftarrow V + \phi_{x_t}\phi_{x_t}^\top
$
and
$
b \leftarrow b + \hat a_{x_t}\phi_{x_t}
$
(Lines~\ref{alg:nrqlin:updateV}--\ref{alg:nrqlin:updateb}).

By combining linear confidence sets with Bayesian noise-aware reward estimation, NR-QLinUCB achieves robustness under realistic NISQ noise while preserving the optimism-driven exploration principle of LinUCB.

\begin{algorithm}[t]
\caption{Noise-Resilient Quantum Linear UCB (NR-QLinUCB)}
\label{alg:NRQLinUCB}
\begin{algorithmic}[1]
\Statex \textbf{Input:} Feature vectors $\{\phi_x\}_{x\in\mathcal{X}}$, time horizon $T$, regularization $\lambda$, confidence level $\delta$
\Statex \textbf{Output:} Arm selection sequence $\{x_t\}_{t=1}^T$

\State Initialize $\hat{a}_x \leftarrow 0,\; s_x \leftarrow 0$ for all $x$
\State $V \leftarrow \lambda I$, $b \leftarrow 0$ \label{alg:nrqlin:initV}
\State $\delta_{\mathrm{r}} \leftarrow \delta/\max(1,T)$
\State $z \leftarrow \sqrt{2 \log\!\left(\frac{T}{\delta_{\mathrm{r}}}\right)}$

\For{$t = 1$ to $T$}
    \State $\hat{\theta} \leftarrow V^{-1} b$
    \label{alg:nrqlin:theta}

    \For{each arm $x\in\mathcal{X}$}
        \label{alg:nrqlin:index-start}
        \State $I_x \leftarrow \phi_x^\top \hat{\theta}
        + \sqrt{\beta_t\, \phi_x^\top V^{-1}\phi_x}
        + z\, s_x$
        \label{alg:nrqlin:index-end}
    \EndFor

    \State $x_t \leftarrow \arg\max_x I_x$
    \label{alg:nrqlin:select}

    \State $(\hat{a}_{x_t}, s_{x_t}) \leftarrow \textbf{BQMC}(\tilde{\mathcal{O}}_{x_t})$
    \label{alg:nrqlin:bqmc}

    \State $V \leftarrow V + \phi_{x_t}\phi_{x_t}^\top$
    \label{alg:nrqlin:updateV}
    \State $b \leftarrow b + \hat{a}_{x_t}\phi_{x_t}$
    \label{alg:nrqlin:updateb}

\EndFor

\State \textbf{return} $\{x_t\}_{t=1}^T$
\end{algorithmic}
\end{algorithm}
 \section{Evaluation}
\subsection{Experimental Setup}

\begin{figure*}[t]
    \centering
    \begin{minipage}{0.32\textwidth}
        \centering
        \includegraphics[width=\linewidth]{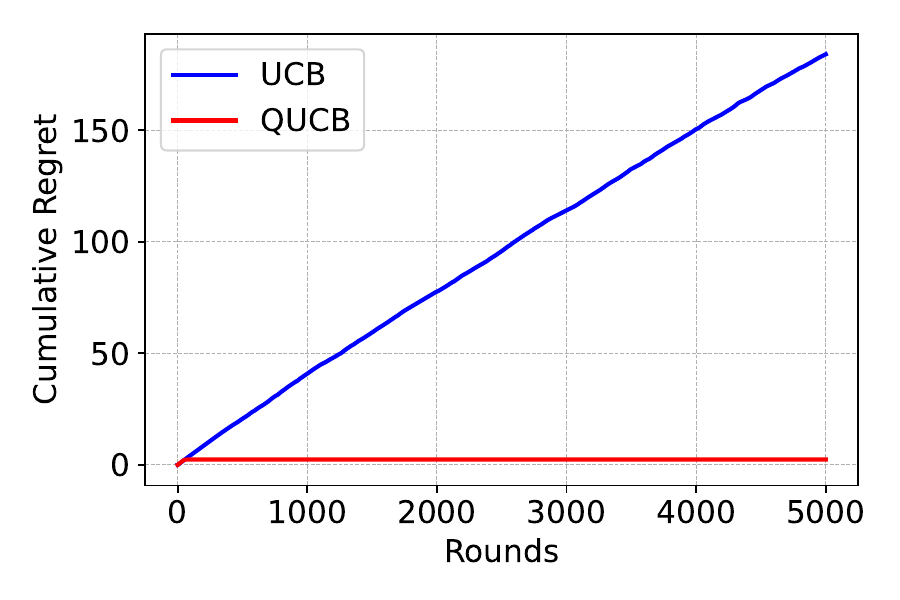}
        \\[-2pt]
        {\footnotesize (a) QUCB vs. UCB with gap $=0.01$}
        \label{fig:qucb-noiseless-1}
    \end{minipage}\hfill
    \begin{minipage}{0.32\textwidth}
        \centering
        \includegraphics[width=\linewidth]{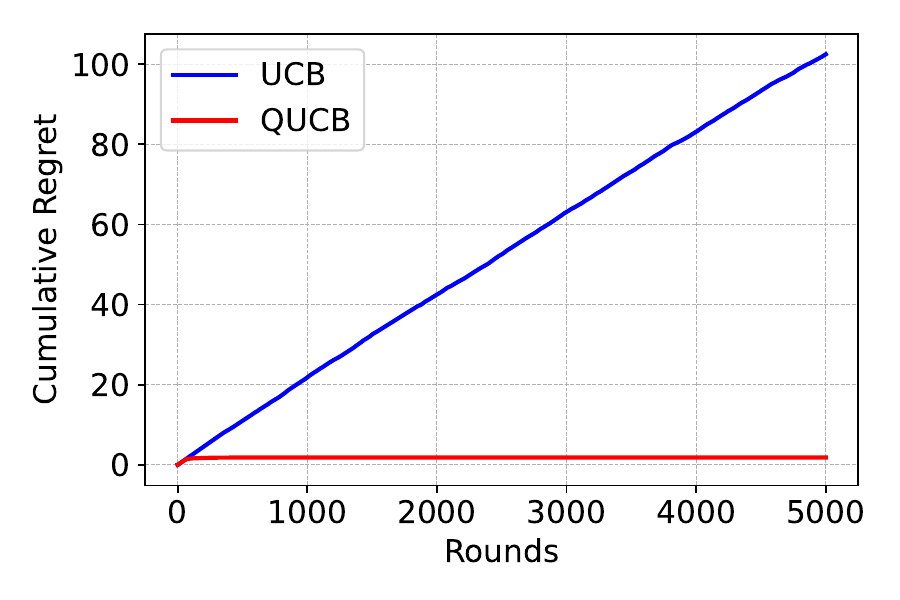}
        \\[-2pt]
        {\footnotesize (b) QUCB vs. UCB with gap $=0.005$}
        \label{fig:qucb-noiseless-2}
    \end{minipage}\hfill
    \begin{minipage}{0.32\textwidth}
        \centering
        \includegraphics[width=\linewidth]{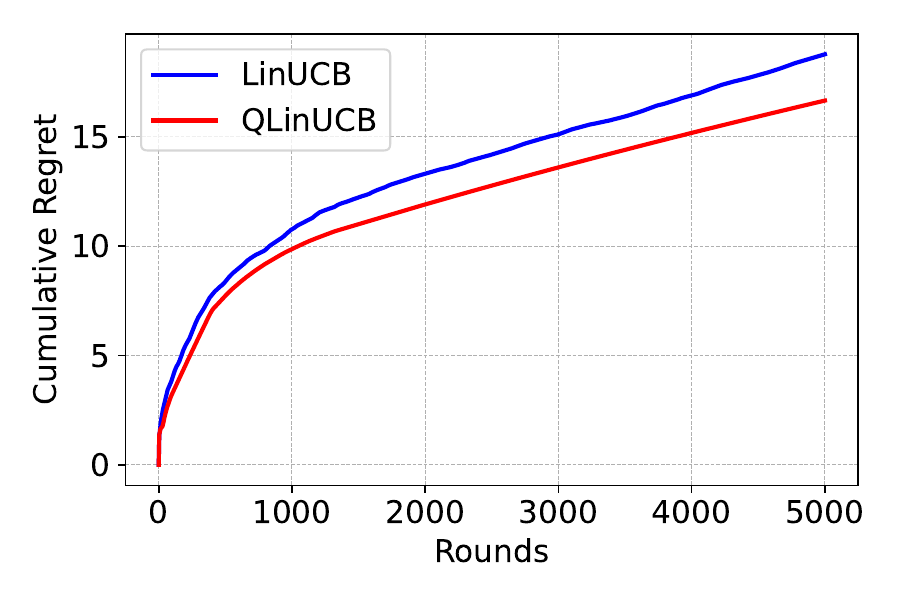}
        \\[-2pt]
        {\footnotesize (c) QLinUCB vs. LinUCB}
        \label{fig:qlinucb-noiseless}
    \end{minipage}
    \caption{Regret comparison results in noiseless settings: (a) and (b) compare QUCB and UCB under two reward gap settings, and (c) compares QLinUCB and LinUCB.}
    \label{fig:noiseless-three-panel}
\end{figure*}

\begin{figure*}[t]
    \centering
    \begin{minipage}{0.24\textwidth}
        \centering
        \includegraphics[width=\linewidth]{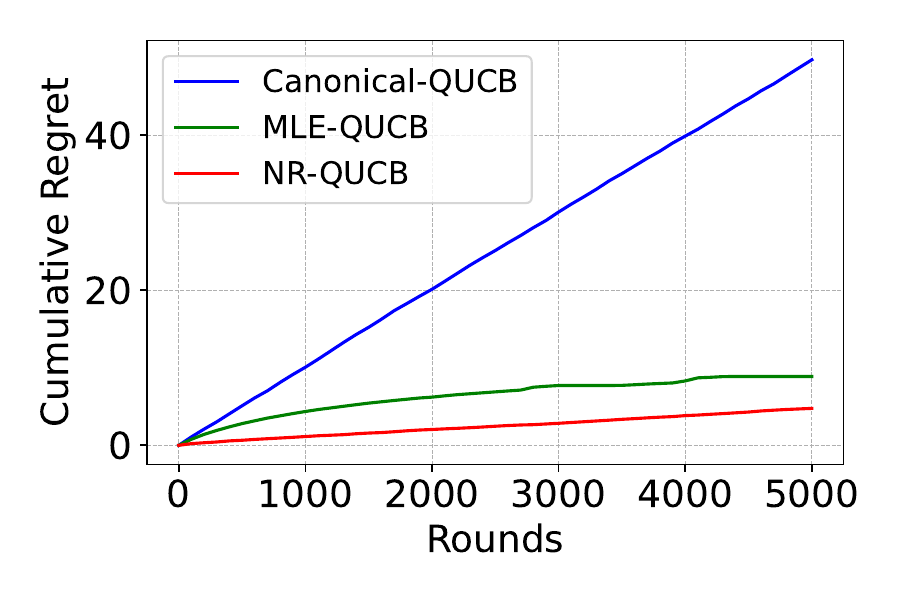}
        \\[-2pt]
        {\footnotesize (a) Exponential decoherence noise}
    \end{minipage}\hfill
    \begin{minipage}{0.24\textwidth}
        \centering
        \includegraphics[width=\linewidth]{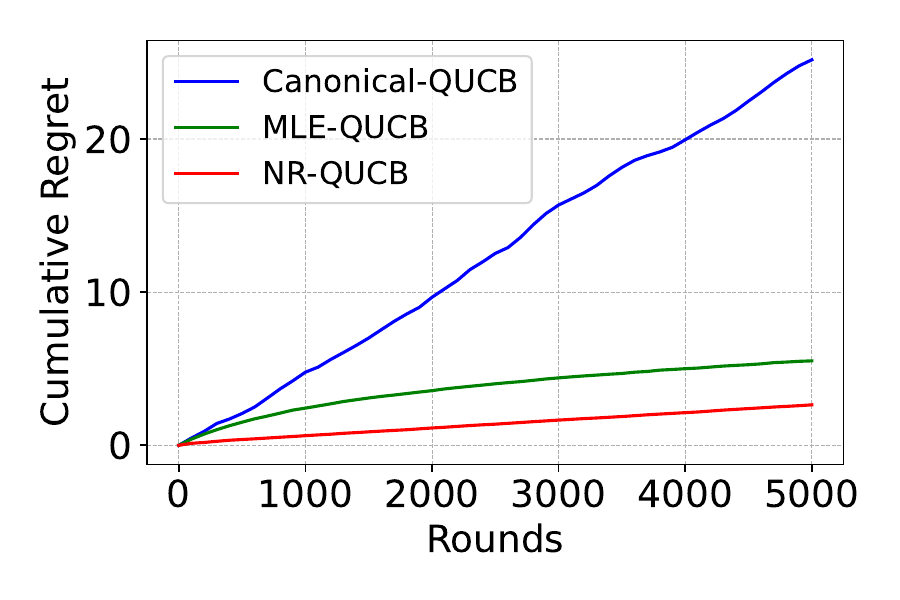}
        \\[-2pt]
        {\footnotesize (b) Readout noise}
    \end{minipage}\hfill
    \begin{minipage}{0.24\textwidth}
        \centering
        \includegraphics[width=\linewidth]{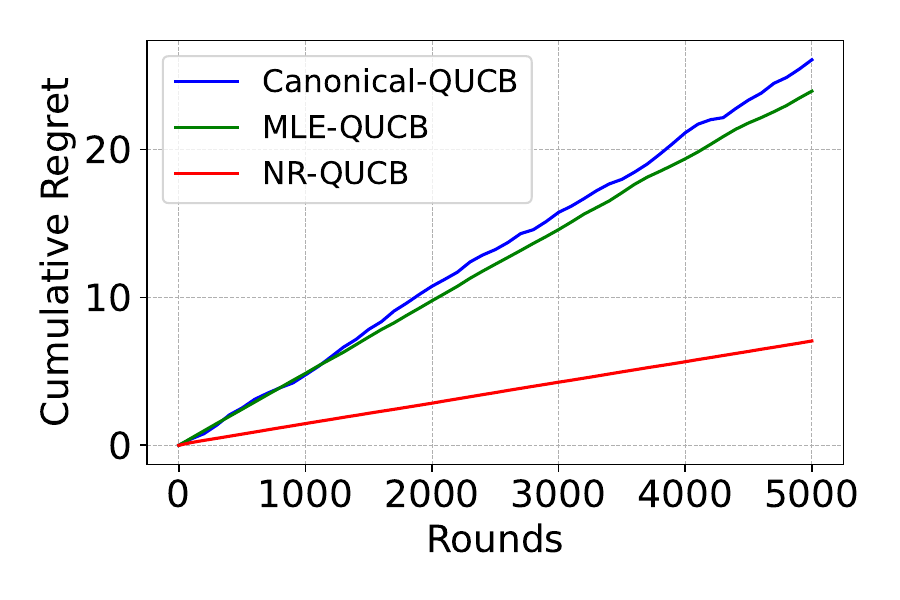}
        \\[-2pt]
        {\footnotesize (c) Depolarizing noise}
    \end{minipage}\hfill
    \begin{minipage}{0.24\textwidth}
        \centering
        \includegraphics[width=\linewidth]{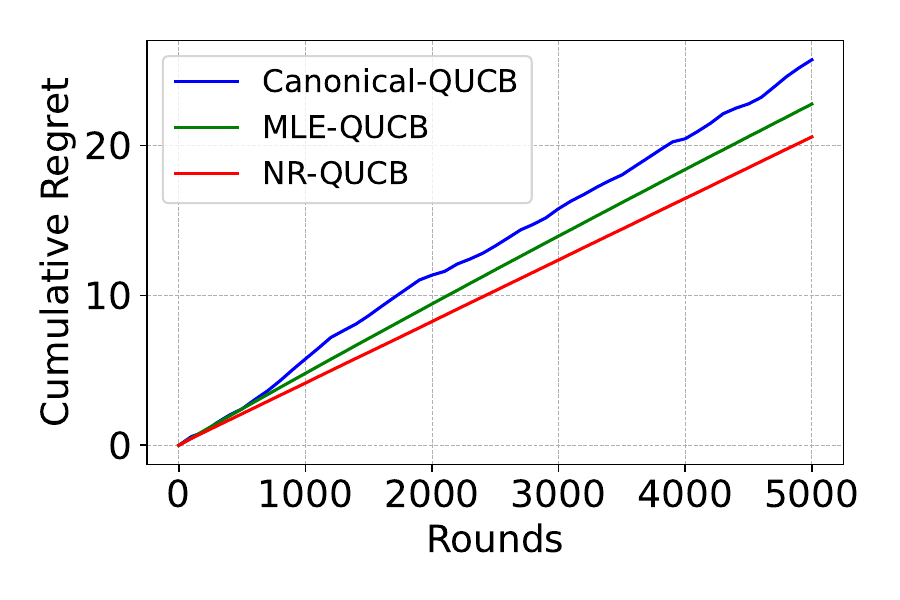}
        \\[-2pt]
        {\footnotesize (d) Amplitude damping noise}
    \end{minipage}
    \caption{Cumulative regret of QUCB variants under four noise settings: exponential decoherence, readout, depolarizing, and amplitude damping.}
    \label{fig:regret-qucb}
\end{figure*}

\begin{figure*}[t]
    \centering
    \begin{minipage}{0.24\textwidth}
        \centering
        \includegraphics[width=\linewidth]{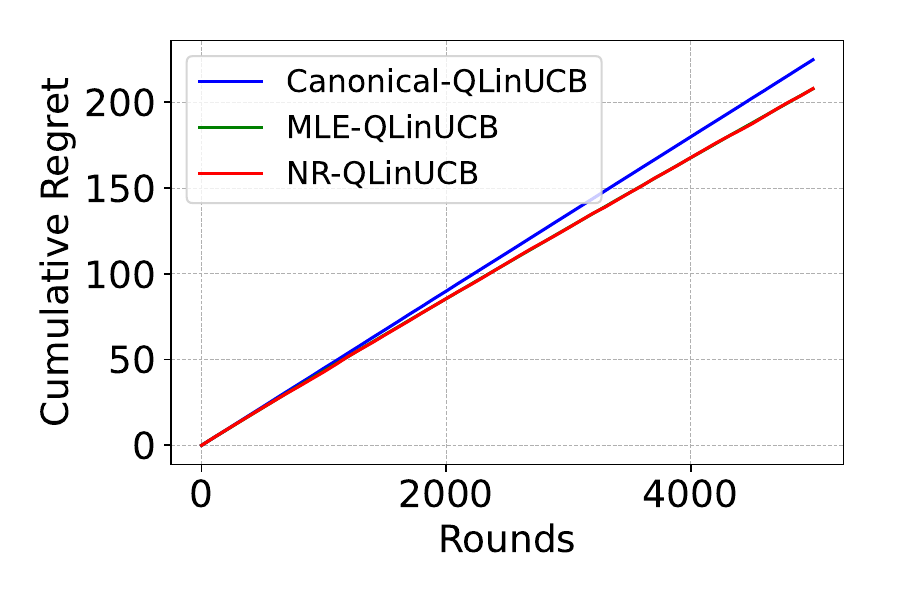}
        \\[-2pt]
        {\footnotesize (a) Exponential decoherence noise}
    \end{minipage}\hfill
    \begin{minipage}{0.24\textwidth}
        \centering
        \includegraphics[width=\linewidth]{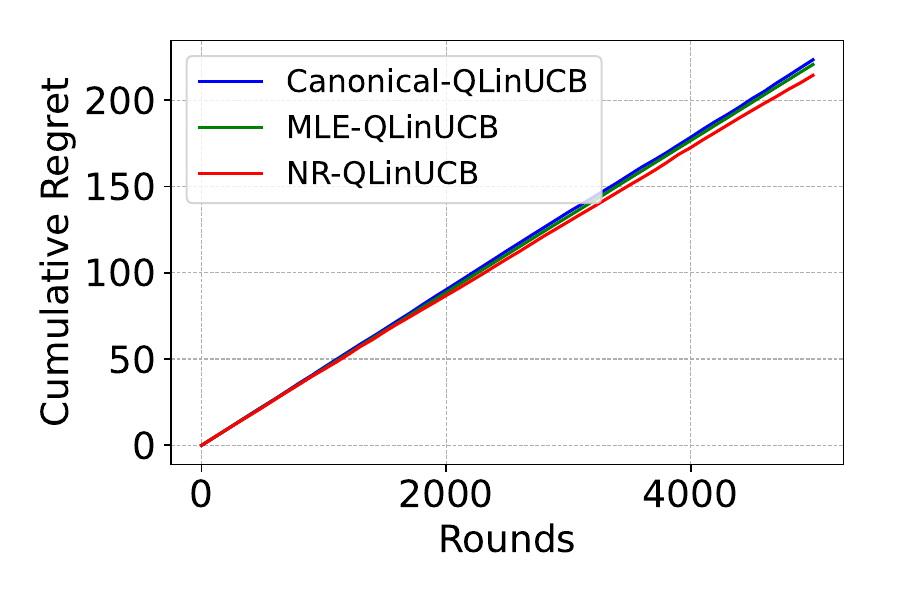}
        \\[-2pt]
        {\footnotesize (b) Readout noise}
    \end{minipage}\hfill
    \begin{minipage}{0.24\textwidth}
        \centering
        \includegraphics[width=\linewidth]{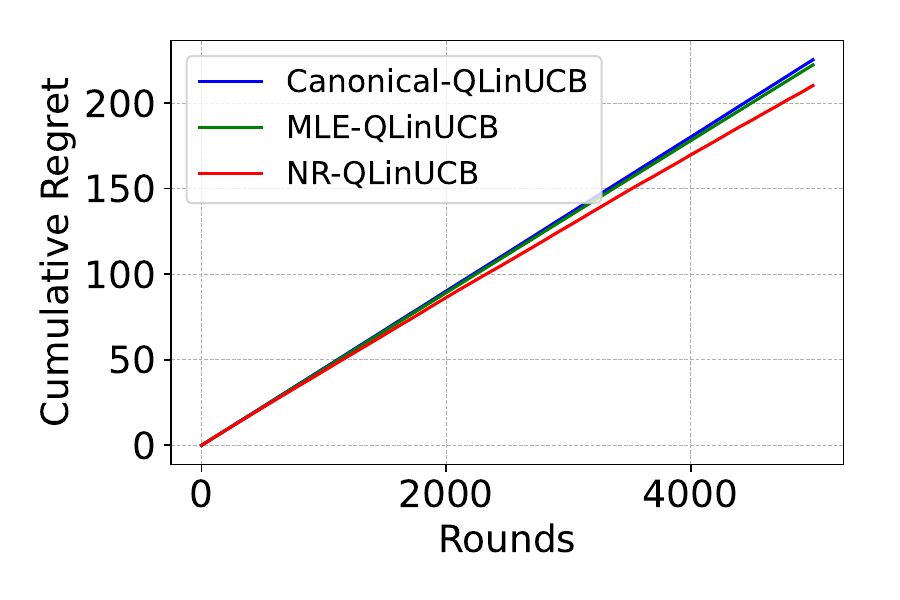}
        \\[-2pt]
        {\footnotesize (c) Depolarizing noise}
    \end{minipage}\hfill
    \begin{minipage}{0.24\textwidth}
        \centering
        \includegraphics[width=\linewidth]{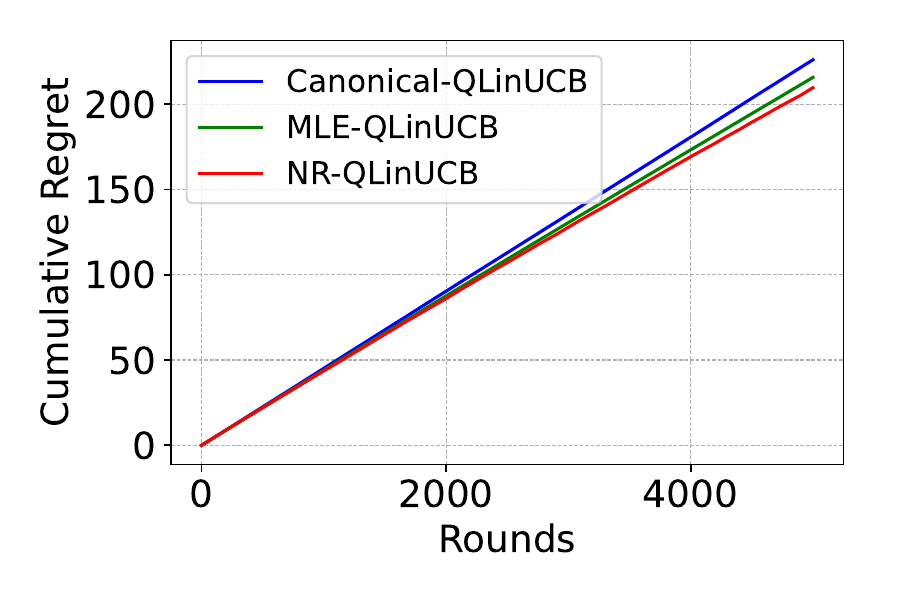}
        \\[-2pt]
        {\footnotesize (d) Amplitude damping noise}
    \end{minipage}
    \caption{Cumulative regret of QLinUCB variants under four noise settings: exponential decoherence, readout, depolarizing, and amplitude damping.}
    \label{fig:regret-qlinucb}
\end{figure*}

We conduct experiments to evaluate the NR-QUCB and NR-QLinUCB under noiseless and noisy conditions. Two quantum amplitude estimation methods are used to estimate reward means:
\begin{itemize}
    \item \emph{Canonical-QAE}: the original phase estimation based method that achieves quadratic speedup but requires deep circuits and QFT.
    \item \emph{MLE-QAE}: a likelihood-based variant that fits measurement outcomes to a statistical model, avoiding QFT and enabling shallower, NISQ-friendly implementations.
\end{itemize}
Accordingly, \emph{Canonical-QUCB} uses Canonical QAE as the QMC subroutine to estimate reward means, while \emph{MLE-QUCB} uses MLE-QAE for mean estimation.

We report results in two parts. Section~\ref{ssec:noiseless} compares regret under ideal (noise-free) QMC: QUCB vs.\ UCB and QLinUCB vs.\ LinUCB. Section~\ref{ssec:noisy} compares NR-QUCB and NR-QLinUCB against the baselines under four noise environments:
\begin{itemize}
    \item \textbf{Exponential Decoherence.} We model decoherence via an effective exponential visibility decay, where the ideal success probability is mixed toward $1/2$ according to $v=\exp(-d/T_c)$, and the circuit depth scale is defined as $d=2m+1$.
    \item \textbf{Readout Noise.} Measurement errors are modeled by applying a bit-flip channel to the measured qubits immediately before measurement. 
    \item \textbf{Depolarizing Noise.} A depolarizing channel is applied to the data qubit after each logical block (after the initial state preparation $A$ and after every Grover block $Q$).
    \item \textbf{Amplitude Damping Noise.} An amplitude damping channel is applied to the data qubit after each logical block. 
\end{itemize}

For all experiments, we repeat each configuration 50 times and report average regret. All numerical simulations were implemented using the MindSpore Quantum framework~\cite{xu2024mindspore}. The source code are publicly available at~\cite{opensource}.

\subsection{Performance in Noiseless Environments}\label{ssec:noiseless}

For QMAB, we set the number of arms to $K=3$, with the optimal arm mean reward fixed at $0.5$. The reward gaps are set to $0.01$ and $0.005$. Each experiment runs for $T=5000$ rounds and is repeated $50$ times; we report the averaged regret. For the classical UCB, we use the UCB variant in~\cite{auer2002finite}.

Figure~1 (a) and Figure~1 (b) compare the regret of QUCB and UCB under two reward-gap settings in noiseless environments. The results show that QUCB achieves substantially lower regret than classical UCB, consistent with the expected quadratic quantum advantage. Moreover, when the reward gap is very small, UCB cannot reliably identify the optimal arm within $5000$ rounds.

For QSLB, we set the number of arms to $K=10$, and the action vectors are placed uniformly on the quarter unit circle in the first quadrant. For $a = 0, \ldots, K-1$, we set $\theta_a = \frac{a}{K-1}\cdot \frac{\pi}{2}$ and $x_a = [\cos\theta_a,\,\sin\theta_a]^\top$, so that $\|x_a\|_2 = 1$. The unknown parameter is $\theta^* = [\cos(\phi\pi/2),\,\sin(\phi\pi/2)]^\top$ with $\phi = 0.7$.

Figure~1 (c) compares the regret of QLinUCB and LinUCB. The results indicate that QLinUCB attains lower regret than LinUCB, mainly because QMC provides more accurate mean estimation.

\subsection{Performance in Noisy Environments}\label{ssec:noisy}
To evaluate the robustness of the algorithms under realistic hardware imperfections, we consider four representative noise models. 
First, we simulate exponential decoherence noise using an exponential visibility decay model with parameter $T_c=2000$.
Second, readout noise are modeled by a bit-flip readout channel with probability $p=0.03$ applied before measurement. 
Third, depolarizing noise is simulated using a depolarizing channel with probability $p=0.05$ inserted after each logical circuit block. 
Finally, amplitude dampling is modeled using an amplitude damping channel with damping parameter $\gamma=0.08$. 

Figure~\ref{fig:regret-qucb} shows the cumulative regret of three QUCB variants under four noise models: Canonical-QUCB, MLE-QUCB, and NR-QUCB.
The results show that BQMC improves mean-reward estimation accuracy in noisy settings, enabling QUCB to identify the best arm more quickly and thereby reducing cumulative regret. Because Canonical-QAE relies on QFT, its deeper circuits accumulate noise more rapidly, which leads to worse estimation performance than MLE-QAE. Consequently, MLE-QUCB achieves lower regret than Canonical-QUCB.

Figure~\ref{fig:regret-qlinucb} shows the cumulative regret of the corresponding QLinUCB variants under the same four noise models: Canonical-QLinUCB, MLE-QLinUCB, and NR-QLinUCB.
The results indicate that NR-QLinUCB still attains the lowest regret. However, the performance gain of BQMC over the other two QMC methods is smaller than in the QUCB setting. This is because the linear model aggregates reward observations, which naturally averages estimation noise over time and reduces the relative benefit of variance-reduction techniques such as BQMC.

\section{Related Works}
\subsection{Quantum Monte Carlo}
Quantum Monte Carlo (QMC) builds on amplitude amplification and estimation. Brassard \emph{et al}. introduced quantum amplitude estimation (QAE), combining Grover-style amplification with quantum phase estimation to achieve a quadratic speedup over classical Monte Carlo: mean estimation error $\epsilon$ costs $O(1/\epsilon)$ queries instead of $O(1/\epsilon^2)$~\cite{brassard2000quantum}. Montanaro generalized this framework to broader Monte Carlo integration, establishing quadratic speedups for expectation estimation under coherent oracle access~\cite{montanaro2015quantum}.

Standard QAE relies on controlled operations and QFT, leading to deep circuits ill-suited for NISQ devices. QFT-free, low-depth variants were therefore proposed, including maximum-likelihood amplitude estimation~\cite{suzuki2020amplitude} and iterative QAE (IQAE)~\cite{grinko2021iterative}, which reduce depth and qubit requirements while retaining near-quadratic speedups.

Noise in NISQ hardware fundamentally alters the statistics of amplitude estimation, introducing bias and variance inflation that can erase the theoretical advantage. Maximum-likelihood amplitude estimation under depolarizing noise jointly estimates amplitude and noise parameters~\cite{tanaka2021amplitude}. Subsequent work incorporated explicit noise models into likelihoods~\cite{herbert2024noise} and developed noise-resilient strategies for gate- and depth-dependent noise~\cite{ding2023general}. Bayesian amplitude estimation has also been extended to noisy devices to improve statistical efficiency and uncertainty quantification~\cite{ramoa2025bayesian}.

\subsection{Quantum Multi-Armed Bandits}
Quantum multi-armed bandit (QMAB) extends classical exploration and exploitation to quantum query models. Early work studied quantum speedups for best-arm identification (BAI)~\cite{casale2020quantum}, showing quadratic improvements in sample complexity under coherent reward oracle access~\cite{wang2021quantum}. These gains rely on amplitude estimation to accelerate mean reward estimation.

Wan \emph{et al}. later proposed quantum algorithms for stochastic multi-armed and stochastic linear bandits under a quantum reward oracle model~\cite{wan2023quantum}. By embedding quantum mean estimation within UCB-type frameworks, they obtained logarithmic regret with improved dependence on the horizon $T$, assuming coherent access to reward-generating mechanisms.

From an information-theoretic view, Lumbreras \emph{et al}. analyzed exploration and exploitation trade-offs for learning quantum states and derived regret lower bounds~\cite{lumbreras2022multi}. Other work examined the oracle model more critically, showing that if arms are general quantum channels, especially when internal classical randomness breaks coherence, quantum query advantages may disappear~\cite{buchholz2025multi}. This underscores that speedups depend fundamentally on the oracle access model.

Overall, QMAB studies largely assume ideal coherent oracles and emphasize asymptotic regret improvements. Robustness under realistic noise, particularly when QMC subroutines incur bias and decoherence, remains underexplored. Bridging noise-aware QMC with regret-optimal bandit strategies is therefore an important open direction.

\section{Conclusion}
TIi ths paper ,swe tudiynoise-resilient quantum bandits by explicitly modeling NISQ noise in quantum Monte Carlo estimation and integrating a Bayesian QMC estimator into multi-armed and stochastic linear bandit aproblem W
e developed NR-QUCB and NR-QLinUCB, which use posterior mean and uncertainty from BQMC to construct robust exploration indices, and analyzed their behavior under realistic noisy oracle access. Experiments under practical noise models demonstrate improved estimation accuracy and regret performance compared with noise-agnostic baselines. Future work includes tighter theoretical guarantees under more general noise processes and validation on hardware implementations.

\section*{Acknowledgment}
This work was sponsored by CPS-Yangtze Delta Region Industrial Innovation Center of Quantum and Information Technology--MindSpore Quantum Open Fund.

\bibliographystyle{IEEEtran}
\bibliography{IEEEabrv,mybibfile}

\end{document}